%% file: iclr2026_conference.tex
\documentclass{article} 
\usepackage{iclr2026_conference,times}

\input{math_commands.tex}

\usepackage{tabularx}
\usepackage{booktabs}
\usepackage{hyperref}
\usepackage{url}
\usepackage{graphicx}
\usepackage{xcolor}
\usepackage{xspace}
\usepackage{booktabs}
\usepackage{pifont}
\usepackage{enumitem}
\usepackage{wrapfig}
\usepackage{lipsum}
\usepackage{caption}

\definecolor{purple}{RGB}{160, 102, 191}
\newtheorem{definition}{\textcolor{black}{Definition}}
\makeatletter
\renewcommand{\@thmcountersep}{\space} 

\makeatother
\newtheorem{proposition}{\textcolor{black}{Proposition}}

\newtheorem{challenge}{\textcolor{black}{Challenge}}


\newcommand\mypar[1]{\par\addvspace{1.0mm}\noindent\textbf{#1}\;}

\title{Streaming Drag-Oriented Interactive Video Manipulation: \textbf{Drag Anything, Anytime}!}

\author{
    Junbao Zhou$^{1}$, Yuan Zhou$^{1}$\thanks{Yuan Zhou (yuan.zhou@ntu.edu.sg) is the corresponding author of the paper.}\,\,\,, 
    Kesen Zhao$^{1}$, Qingshan Xu$^{2}$, Beier Zhu$^{2}$, Richang Hong$^{3}$,\\
    \textbf{ Hanwang Zhang$^{1}$} \\
    $^{1}$Nanyang Technological University \qquad $^{2}$University of Science and Technology of China \\
    $^{3}$Hefei University of Technology \\
    \small\texttt{\{yuan.zhou, hanwang.zhang\}@ntu.edu.sg} \\
    \small\texttt{\{qingshan.xu, beier.zhu\}@ustc.edu.cn} \\
    \small\texttt{\{JUNBAO001, KESEN002\}@e.ntu.edu.sg, }\small\texttt{hongrc.hfut@gmail.com}
}

\iclrfinalcopy 
\begin{document}
\maketitle

\vspace{-0.8cm}

\begin{center}
Code: \url{https://github.com/junbao-zhou/DragStream}
\end{center}

\begin{center}
Project page: \url{https://junbao-zhou.github.io/DragStream.github.io/}
\end{center}

\begin{figure}[h]
    \centering
    \includegraphics[width=1\linewidth]{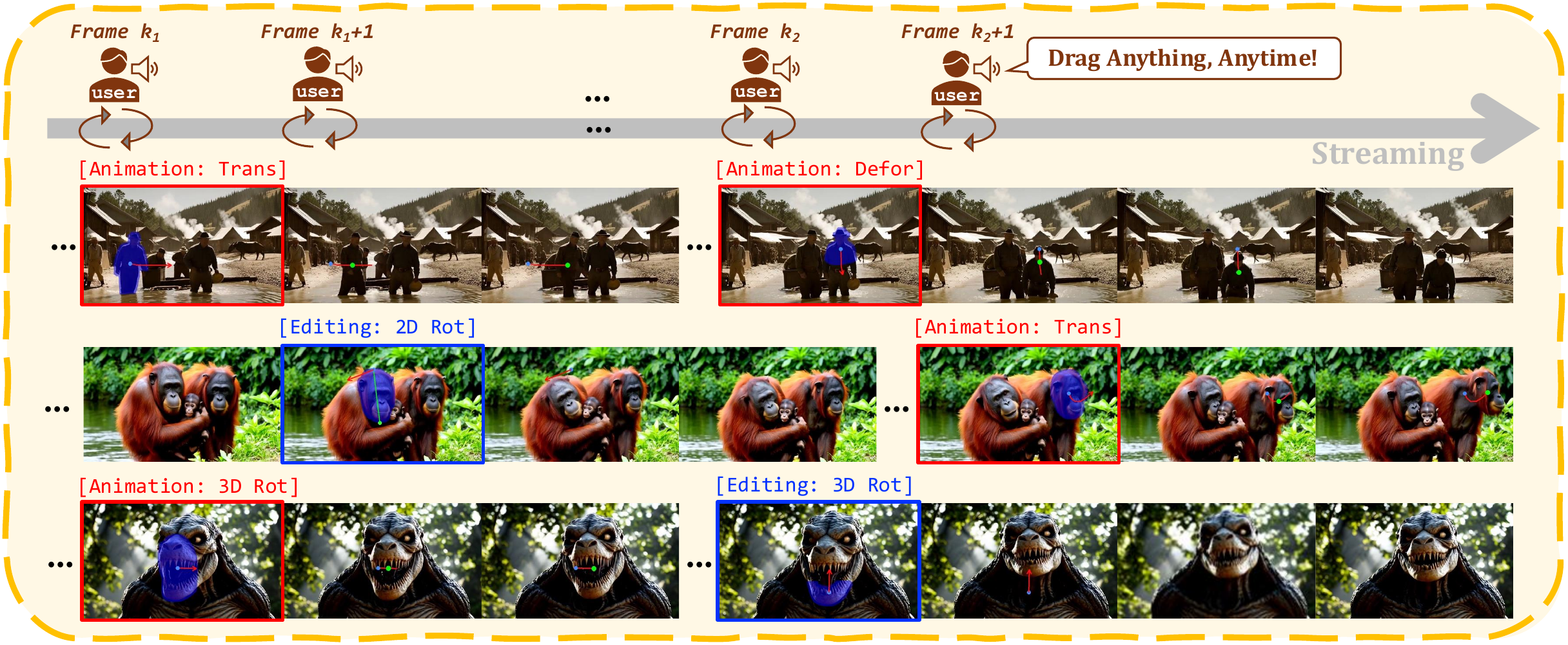}
    \vspace{-0.6cm}
    \caption{\textbf{Examples of our REVEL task.} The streaming video manipulation results shown above---including both \texttt{Editing} and \texttt{Animation}  with drag effects such as object translation (``\texttt{Trans}''), deformation (``\texttt{Defor}''), and rotation (``\texttt{Rot}'')---are produced by our \textbf{DragStream} method.}
    \label{fig:1}
\end{figure}

\begin{abstract}
 Achieving streaming, fine-grained control over the outputs of autoregressive video diffusion models remains challenging, making it difficult to ensure that they consistently align with user expectations. To bridge this gap, we propose \textbf{stReaming drag-oriEnted interactiVe vidEo manipuLation (REVEL)}, a new task that enables users to modify generated videos \emph{anytime} on \emph{anything} via fine-grained, interactive drag. Beyond DragVideo and SG-I2V, REVEL unifies drag-style video manipulation as editing and animating video frames with both supporting user-specified translation, deformation, and rotation effects, making drag operations versatile. In resolving REVEL, we observe: \emph{i}) drag-induced perturbations accumulate in latent space, causing severe latent distribution drift that halts the drag process; \emph{ii}) streaming drag is easily disturbed by context frames, thereby yielding visually unnatural outcomes. We thus propose a training-free approach, \textbf{DragStream}, comprising: \emph{i}) an adaptive distribution self-rectification strategy that leverages neighboring frames' statistics to effectively constrain the drift of latent embeddings; \emph{ii}) a spatial-frequency selective optimization mechanism, allowing the model to fully exploit contextual information while mitigating its interference via selectively propagating visual cues along generation. Our method can be seamlessly integrated into existing autoregressive video diffusion models, and extensive experiments firmly demonstrate the effectiveness of our DragStream.



\end{abstract}

\section{Introduction}
\label{sec:intro}
\vspace{-0.3cm}

Video Diffusion Models (VDMs) have shown impressive capabilities in generating photorealistic videos, and their success inspired a broad range of generative applications, including image animation \cite{animateanything,animateanyone}, text-based video editing \cite{pix2video,videop2p}, camera-controlled video generation \cite{cami2v,cameractrl,recammaster}, etc. With the recent remarkable progress in autoregressive VDMs \cite{causvid,selfforcing} and diffusion acceleration \cite{zhu2025enhancing, wang2025paralleldiffusionsolverresidual}, researchers have been focusing more on achieving controllable video generation in a streaming manner, thereby enabling users to interact with VDMs and alter synthetic videos on the fly. For instance, \cite{streamdiffusion,streamdit,autoregap} proposed directly finetuning VDMs to support streaming video generation conditioned on text, camera viewpoint, and human pose, whereas \cite{Streamv2v} realized training-free, text-guided streaming video translation by introducing a looking-back strategy. 

Drag-style operations have become a crucial control signal for VDMs due to their fine-grained nature and user-friendly interactivity \cite{dragnext,draganything,dragvideo,motionctrl,sgi2v,realtimemoviediff}. However, it remains challenging to realize streaming, fine-grained control over the outputs of VDMs through drag-style operations. To mitigate this dilemma, we propose a new task, \textbf{stReaming drag-oriEnted interactiVe vidEo manipuLation (REVEL)}. As shown in Figure \ref{fig:1}, REVEL aims to allow users to modify generated videos \emph{at any time} and \emph{on any content} via fine-grained, interactive drag, making generated videos consistently meet users' requirements. We go beyond prior methods, such as DragVideo \cite{dragvideo} and SG-I2V \cite{sgi2v}, by unifying drag-oriented video manipulation as editing and animating video frames, with both supporting user-specified translation, deformation, and rotation effects, thereby making drag operations versatile and establishing a standard paradigm for drag-style video manipulation.

Given the fine-grained nature and high diversity of drag-based video manipulation, solving REVEL is non-trivial. Directly finetuning VDMs to realize REVEL usually incurs expensive training costs---requiring training VDMs on large-scale, fine-grained drag-style data by hundreds or even thousands of H100 GPU hours \cite{causvid,streamdit,selfforcing}---making it impractical for resource-constrained scenarios. \textbf{This observation naturally leads us to ask a key question: \emph{How can high-quality REVEL be achieved without incurring prohibitive computational costs?}} 

We propose solving the above question from a training-free perspective in this paper, so as to effectively reduce training expenses. However, we observe that there exist two key challenges: \emph{i}) perturbations induced by drag operations easily accumulate in latent space, thereby causing severe latent distribution drift that totally halts the drag process; \emph{ii}) streaming drag is easily disturbed by context frames, resulting in visually unnatural content. Therefore, we propose a new \textbf{DragStream} approach. Specifically, we first design an Adaptive Distribution Self-Rectification (ADSR) strategy that suppresses the distribution drift of latent code by considering statistics from neighboring frames, thereby effectively overcoming drag interruption. We also introduce a Spatial-Frequency Selective Optimization (SFSO) mechanism, which propagates visual cues from preceding video frames selectively in both spatial and frequency domains. As a result, we can fully exploit the information of context frames while relieving their interference. ADSR and SFSO enable our DragStream to achieve high-quality results on REVEL without incurring prohibitive training costs, while allowing it to be seamlessly integrated into existing autoregressive VDMs. Extensive experiments provided in Section \ref{sec:exp} and the appendix consistently demonstrate the superiority of our proposed approach.

Here, we summarize the main contributions of this paper:
\vspace{-0.2cm}
\begin{itemize}[leftmargin=3.5em]
    \item We propose \textbf{stReaming drag-oriEnted interactiVe vidEo manipuLation (REVEL)}, a new task that enables users to drag \emph{anything anytime} during video generation, thus achieving streaming, fine-grained control over the outputs of VDMs via drag-style operations.
    \item We identify two key challenges in solving REVEL within a training-free paradigm: \emph{i}) drag-induced perturbations cause severe latent distribution drift and halt the drag process; and \emph{ii}) streaming drag is disturbed by context frames, resulting in visually unnatural outcomes.  
    \item We propose \textbf{DragStream}, which incorporates a Spatial-Frequency Selective Optimization (SFSO) mechanism and an Adaptive Distribution Self-Rectification (ADSR) strategy to effectively suppress context interference and mitigate distribution drift in latent code.
    \item Extensive experiments clearly demonstrate the effectiveness of our approach in addressing REVEL, showing that it achieves high-quality streaming drag-style manipulation, remains training-free, and offers plug-and-play integration with existing autoregressive VDMs.
\end{itemize}

\section{Related Work}
\label{sec:rela}
\vspace{-0.3cm}
%

\textbf{Streaming Video Generation.} StreamDiffusion \cite{streamdiffusion}, SVDiff \cite{streamingvd}, and StreamDiT \cite{streamdit} are recent representative streaming text-guided video generation models, in which VDMs are either trained from scratch or finetuned to enable streaming control via text prompts. \cite{autoregap} proposed an autoregressive adversarial post-training strategy that enables VDMs to operate as one-step autoregressive generators, supporting conditions on human pose, camera viewpoint, and text. 
\cite{Streamv2v} designed a text-based streaming video translation model by preserving historical information across video frames using a feature bank. 

\textbf{Drag-Based Video Generation and Editing.} 
\cite{draganything,motionctrl} proposed finetuning bidirectional VDMs with trajectory conditions, thereby realizing trajectory-guided video generation. \cite{tora} proposed unifying text, image, and trajectory conditions into a DiT framework \cite{dit}, while \cite{motionprompt,motionpro} further trained VDMs on dense trajectories. \cite{sgi2v,freetraj,peekaboo,dragvideo} resorted to training-free frameworks. \cite{dragvideo} introduced a drag-based latent optimization strategy to realize drag-oriented video editing. \cite{sgi2v} further considered semantically aligned features \cite{zhu2025hierarchical,zhu2024selective,zhu2025enhancing} during dragging, whereas \cite{freetraj} achieved trajectory-guided video generation by imposing guidance on both attention and noise construction. 

\textbf{REMARK 1.} i) Despite the progress in streaming video generation, current models rarely support highly flexible, fine-grained drag-style operations in a streaming manner---a key challenge our work aims to address. ii) Existing drag-based video generation and editing methods are not tailored for streaming tasks, making them unsuitable for achieving fine-grained, streaming  control over the outputs of autoregressive VDMs. iii) Directly finetuning VDMs for realizing streaming drag-style manipulation is computationally expensive, usually  requiring training VDMs on large-scale drag-style data by hundreds or even thousands of H100 GPU hours, which is unacceptable for resource-constrained scenarios. Different from finetuning-based methods, our DragStream is training-free and can be seamlessly integrated into existing autoregressive VDMs. iv) Beyond previous works, we unify drag-style video manipulation as editing and animating video frames with both supporting user-specified translation, deformation, and rotation effects, thus making drag operations versatile.

\vspace{-0.3cm}

\section{Streaming Drag-Oriented Interactive Video Manipulation}
\label{sec:revel}
We first give the definition of our \textbf{stReaming drag-oriEnted interactiVe vidEo manipuLation (REVEL)} task in \textbf{Definition \ref{def:revel}}. For the summary of the main notations, please refer to Section~\ref{sec:notion}.
\begin{definition}[REVEL]
\label{def:revel}
Let $\bm{\Gamma}_k$ denote the $k$-th video frame produced by autoregressive VDMs. REVEL aims to enable users to utilize drag-style operations $\bm{U}_k$ to modify video frames for $\forall k \in \mathbb{Z}^{+}$ and ensures that subsequently nearby frames are consistent to $\bm{\Gamma}_k$, so as to realize streaming, fine-grained control over outputs of VDMs and make generated videos always meet users' requirements.
\end{definition}

We argue that there exist a major limitation in current drag-based video manipulation, namely the lack of a unified definition of drag-style manipulation operations. Existing drag-based video editing methods focus on dragging objects in generated videos, with the goal of yielding the effects of \emph{translation}, \emph{deformation}, and \emph{rotation} \cite{dragvideo}; also, these methods are generally unable to allow users to animate video frames via dragging. By contrast, trajectory-guided video generation models are designed to generate video clips by moving objects along trajectories, with their motion rendered by VDMs; however, they are not flexible enough to specifically allow users to determine the type of drag operations, e.g., deforming object shape, translating objects, or rotating them around a center point  \cite{sgi2v,tora}. Since both of these settings are incomplete, we propose unifying drag-style video manipulation operations in \textbf{Proposition \ref{prop:dea}}.

\begin{proposition}[Unifying Drag-Style Video Manipulation Operations]
\label{prop:dea}
We unify drag-style video manipulation as enabling users to perform editing and animation on video frames via drag-style operations, with both supporting user-specified translation, deformation, and 2D/3D rotation effects. Here, editing refers to directly modifying the content of generated video frames, whereas animation represents generating a video clip from an existing frame according to user-given drag instructions.
\end{proposition}

\textbf{REMARK 2.} Here, we clarify how our REVEL task differs from prior works on drag-based video editing and generation. DragVideo \cite{dragvideo} is a recent typical drag-based video editing approach. Different from our REVEL, it only supports drag-based editing and does not allow users to animate video frames. Moreover, DragVideo does not support the 2D object rotation operation. SG-I2V \cite{sgi2v} and Tora \cite{tora} are two typical trajectory-guided video generation approaches. Both of them focus solely on animating images by moving objects along trajectories with VDM-rendered motion, without allowing users to flexibly achieve more fine-grained drag-style effects, such as editing object shape or rotating objects around a center point by a specific angle. Also DragNeXt \cite{dragnext} does not support streaming-style editing. Most importantly, these methods are all incapable of achievinbg drag-oriented video editing and animation in a streaming manner.

We propose addressing REVEL from a training-free perspective, and identify that there exist two key challenges, summarized in \textbf{Challenge~\ref{chal:1}} and \textbf{Challenge~\ref{chal:2}}, respectively.

\begin{figure}[t!]
    \centering
    \includegraphics[width=0.9\linewidth]{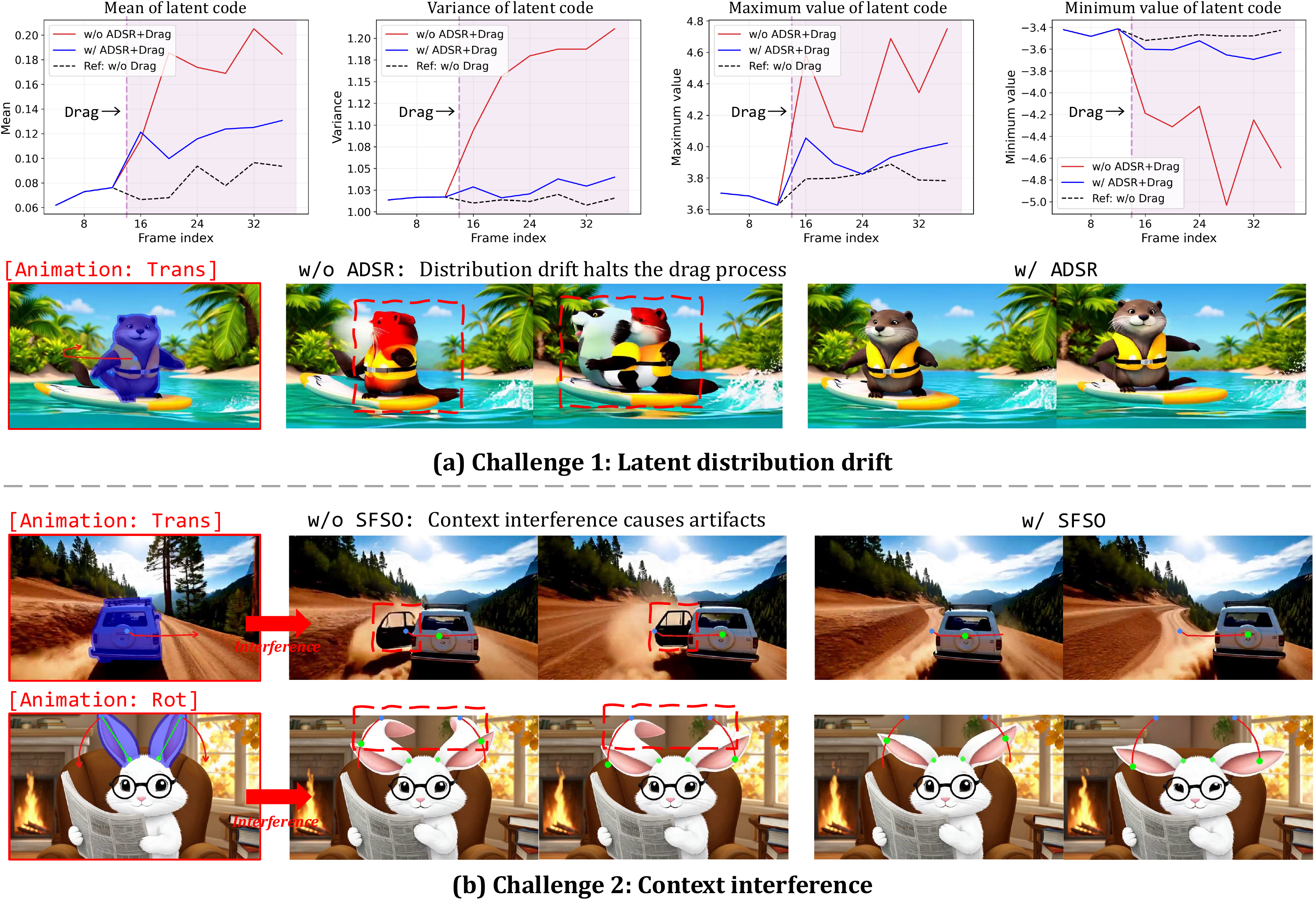}
   \vspace{-0.1cm}
    \caption{\textbf{Examples of Challenge \ref{chal:1} and Challenge \ref{chal:2}.}}
    \label{fig:2}
    \vspace{-0.6cm}
\end{figure}

\begin{challenge}[Latent Distribution Drift]
\label{chal:1}
Perturbations induced by drag-style operations easily accumulate in the latent space of autoregressive VDMs, which leads to severe distribution drift of latent code and thus interrupts the drag process.
\end{challenge}
We show \textbf{Challenge~\ref{chal:1}} in Figure \ref{fig:2} (a). The figure shows that the mean and variance of latent embeddings change significantly once drag operations are applied, while the maximum and minimum values exhibit obvious fluctuations. This instability drives the latent embeddings (``w/o ADSR+drag'') to drift away from the original distribution (``Ref: w/o Drag’’), thereby disrupting the drag process. We find that latent distribution drift may cause undesirable change of object attributes, such as color and category, as shown in the second row of Figure \ref{fig:2} (a). The use of our ADSR strategy (``w/ ADSR+Drag'') can effectively suppress the distribution drift. We will introduce it in Section \ref{sec:adsr}.

\begin{challenge}[Context Interference]
\label{chal:2}
Streaming drag is easily disturbed by context frames, misleading VDMs to produce visually unnatural content and thus substantially degrading video quality.
\end{challenge}
We show \textbf{Challenge~\ref{chal:2}} in Figure \ref{fig:2} (b). The results in Figure \ref{fig:2} (b) clearly indicate that visual cues from previous frames may mislead the subsequent generation severely, e.g., the features around the handle points spuriously guide the model to produce duplicated ears on the rabbit and artifacts on the car (``w/o SFSO''), which obviously lowers the quality of generated videos. We will introduce how to overcome context interference by using our SFSO strategy in Section \ref{sec:sfso}. 
\begin{figure}[t!]
    \centering
    \includegraphics[width=0.95\linewidth]{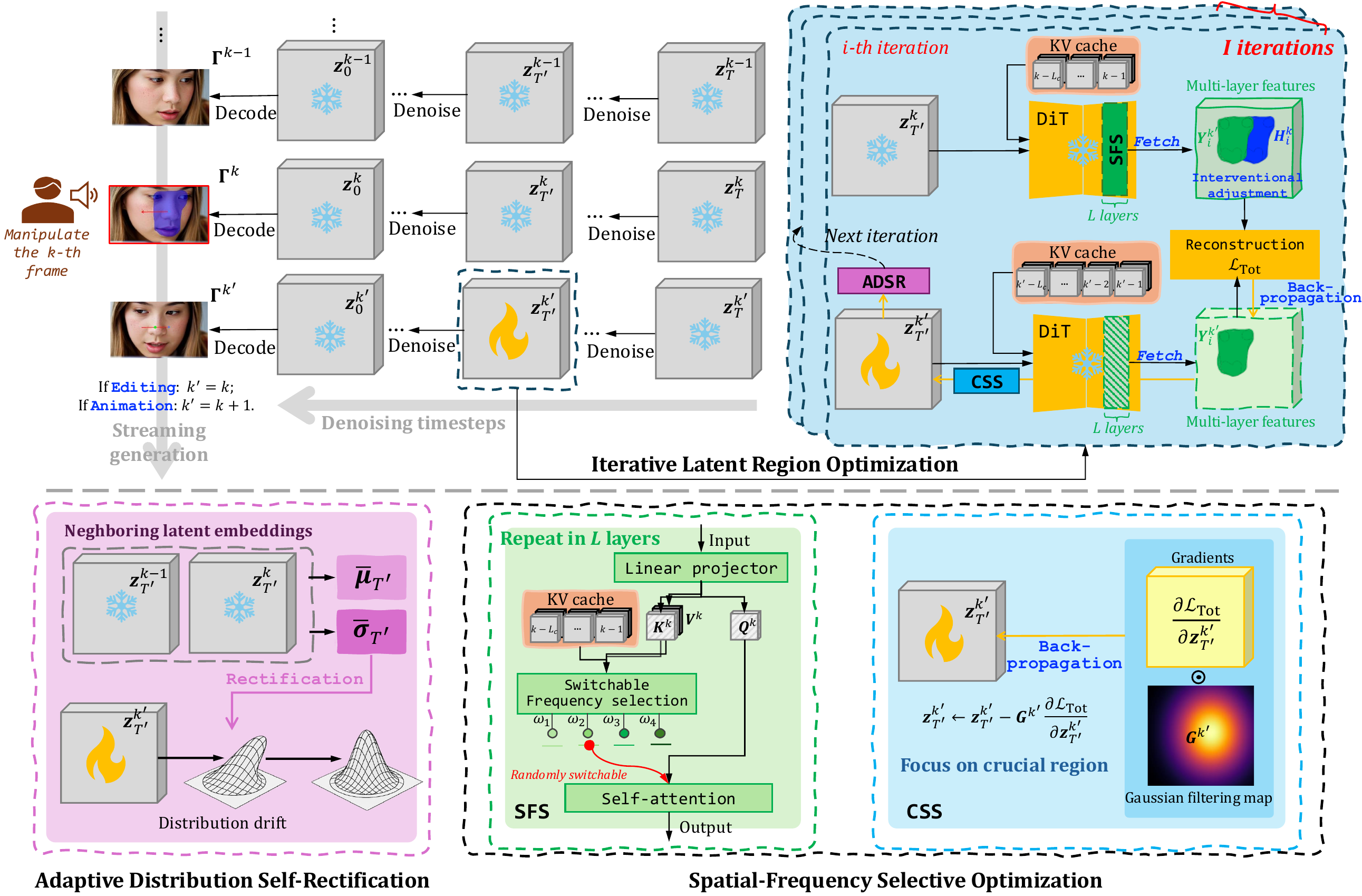}
   \vspace{-0.2cm}
    \caption{\textbf{Schematic illustration of our DragStream,} where an Adaptive Distribution Self-Rectification (ADSR) strategy and a Spatial-Frequency Selective Optimization (SFSO) mechanism are designed to suppress latent distribution drift and context interference, respectively.  }
    \vspace{-0.4cm}
    \label{fig:3}
\end{figure}
\vspace{-0.4cm}

\section{Methodology}
\vspace{-0.3cm}
\subsection{Preliminaries}
\vspace{-0.2cm}
\label{sec:prel}

\textbf{Autoregressive Video Diffusion Models.} Autoregressive VDMs refer to a hybrid generative framework that integrates diffusion models with chain-rule decomposition, i.e.,
$\mathbb{P}(\bm{\Gamma}^{1:k}) = \prod_{i=1}^k \mathbb{P}(\bm{\Gamma}^i \mid \{\bm{\Gamma}^j\}_{j=0:i-1})$, where $\mathbb{P}(\bm{\Gamma}^i \mid \{\bm{\Gamma}^j\}_{j=0:i-1})$ is modeled by iteratively denoising a Gaussian noise latent code $\bm{z}_T^i\in\mathcal{N}(\bm{0},\bm{I})$ to a clean latent code $\bm{z}_0^i$
conditioned on proceeding frames $\{\bm{\Gamma}^j\}_{j=0:i-1}$. The KV caching strategy \cite{Rmem,selfforcing,causvid} is often employed during inference to further reduce computations, thereby accelerating autoregressive generation.

\textbf{Drag-Style Operation Formats.} Following \cite{dragnext}, we use $\bm{U}^k=\{\bm{E}^k,\bm{C}^k\}$ to represent drag operations for $\bm{\Gamma}^k$, where $\bm{E}^k=\{\bm{H}_i^k\}_{i=1:n}$ indicates a set of user-specified handle regions that require to be dragged, and $\bm{C}^k=\{\eta^k,\zeta_i^k,\bm{O}_i^k\}_{i=1:n}$ represents the corresponding drag instructions. The indictor $\eta^k=\texttt{Editing} \text{ or } \texttt{Animation}$ determines whether the video frame $\bm{\Gamma}^k$ is to be edited or animated, whereas $\zeta_i^k$ indicates the type of each drag operation. For \text{animation}, $\bm{O}_i^k=\{\bm{h}_{i}^k,\{\bm{p}_{i}^{k'}\}_{k'=k+1:k+m},\bm{c}_i^k\}$, if $\zeta_i^k=\texttt{Rotation}$; otherwise, $\bm{O}_i^k=\{\bm{h}_{i}^k,\{\bm{p}_{i}^{k'}\}_{k'=k+1:k+m} \}$. Here, $\bm{h}_{i}^k$ represents a handle point, $\{\bm{p}_{i}^{k'}\}_{k'=k+1:k+m}$ represents $m$ discrete target points sampled along a drag trajectory, assigned to subsequent $m$ video frames, and $\bm{c}_i^k$ denotes a rotation center of the handle region $\bm{H}_i^k$. For \text{editing}, $\bm{O}_i^k=\{\bm{h}_{i}^k,\bm{p}_{i}^{k},\bm{c}_i^k\}$, if $\zeta_i^k=\texttt{Rotation}$; otherwise, $\bm{O}_i^k=\{\bm{h}_{i}^k, \bm{p}_{i}^k\}$. Here, each drag operation considers only one target point, since the editing task ignores intermediate drag states. Also, a binary mask $\bm{M}^k$ is utilized to specify the non-editable region of the frame $\bm{\Gamma}^k$.

\subsection{DragStream: Drag Anything, Anytime in a Training-Free Paradigm}
\subsubsection{Overall Pipeline}
\label{sec:pipe}
We first introduce the overall pipeline of our DragStream. Suppose that users observe the video frame $\bm{\Gamma}^{k}$ during streaming generation and intend to manipulate $\bm{\Gamma}^{k}$ by giving the instructions $\bm{U}^k=\{\bm{E}^k,\bm{C}^k\}$, where $\bm{E}^k=\{\bm{H}_i^k\}_{i=1:n}$ indicates handle regions, and $\bm{C}^k=\{\eta^k,\zeta_i^k,\bm{O}_i^k\}_{i=1:n}$ denotes the corresponding drag instructions. We use $\bm{\Gamma}^{k’}$ to represent a video frame produced during dragging, where $k'=k$ if $\eta^k=\texttt{Editing}$; otherwise, $k'>k$ since new frames are animated during \texttt{Animation}. 

We take the handle region $\bm{H}^k_i$ as an example to illustrate our method. As exhibited in Figure \ref{fig:3}, we first denoise $\bm{z}^{k'}_T$ to $\bm{z}^{k'}_{T'}$, and extract the features $\mathcal{F}(\bm{z}^{k'}_{T'})$ by concatenating features from the multiple layers of the DiT denoiser $\epsilon_{\bm{\Theta}}(\cdot|\{\bm{K}^{i},\bm{V}^{i}\}_{i=0:k'-1})$, where $\{\bm{K}^{i},\bm{V}^{i}\}_{i=0:k'-1}$ are the cached keys and values of context frames. We then estimate the position of the handle region $\bm{H}_i^k$ after being dragged within the features $\mathcal{F}(\bm{z}_{T'}^{k'})$ according to the user-given drag instruction:
\begin{equation}
\bm{Y}_{i}^{k'},\bm{\Pi}_{\bm{H}_i^k\rightarrow\bm{Y}_{i}^{k'}}=\mathcal{G} (k',\bm{H}_i^k, \eta^k,\zeta_i^k,\bm{O}_i^k)\text{,}
\label{eq:1}
\end{equation}
\begin{equation*}
s.t., \mathcal{G} (k',\bm{H}_i^k, \eta^k,\zeta_i^k,\bm{O}_i^k)=
\begin{cases}
  \texttt{Rot}(\bm{H}_i^k,\bm{c}_i^k,\theta=\angle\bm{p}_{i}^{k'}\bm{c}_i^k\bm{p}_{i}^k), & \text{if} \text{ $\zeta_i^k=\texttt{Rotation}$} \\
  \texttt{Trans}(\bm{H}_i^k,\bm{\vartheta}=\bm{p}_{i}^{k'}-\bm{p}_{i}^k), & \text{else.}
\end{cases}
\end{equation*}
Here, $\texttt{Rot}(\bm{H}_i^k,\bm{c}_i^k,\theta)$ denotes rotating the handle region $\bm{H}_i^k$ around the center point $\bm{c}_i^k$ by an angle $\theta$, and $\texttt{Trans}(\bm{H}_i^k,\bm{\vartheta})$ indicates translating $\bm{H}_i^k$ by an offset $\bm{\vartheta}$. $\bm{Y}_{i}^{k'}$ is a binary mask that indicates the target position of $\bm{H}^k_i$ in the extracted features $\mathcal{F}(\bm{z}_{T'}^{k'})$, and $\bm{\Pi}_{\bm{H}_i^k\rightarrow\bm{Y}_{i}^{k'}}$ is the coordinate mapping from $\bm{H}_i^k$ to $\bm{Y}_{i}^{k'}$. Finally, the latent code $\bm{z}^{k'}_{T'}$ is iteratively optimized. In each iteration, the features of $\bm{z}^{k}_{T'}$ are also extracted and detached as reference features, $\mathcal{F}_{\texttt{ref}}(\bm{z}^{k}_{T'})=\mathcal{F}(\bm{z}^{k}_{T'}).\texttt{detach()}$. 
Moreover, we interventionally adjust the reference features according to the coordinate mapping, $\mathcal{F}_{\texttt{ref}}(\bm{z}_{T'}^k)[\bm{\Pi}_{\bm{H}_i^k\rightarrow\bm{Y}_{i}^{k'}}]$, thereby perturbing the original latent code and transforming the handle region features to the target position $\bm{Y}_{i}^{k'}$. The latent code $\bm{z}^{k'}_{T'}$ of the new frame $\bm{\Gamma}^{k'}$ can be updated by reconstructing the features from the original handle region at the target position of $\mathcal{F}(\bm{z}_{T'}^{k'})$
\begin{equation}
  \bm{z}_{T'}^{k'} \longleftarrow\bm{z}_{T'}^{k'} - \frac{\partial\mathcal{L}_{\texttt{Tot}}}{\partial\bm{z}_{T'}^{k'}}\text{,}
\label{eq:2}
\end{equation}
where
\begin{equation}
\mathcal{L}_{\texttt{Tot}}=\underbrace{\|\mathcal{F}(\bm{z}_{T'}^{k'})*\bm{Y}_{i}^{k'}-\mathcal{F}_{\texttt{ref}}(\bm{z}_{T'}^k)[\bm{\Pi}_{\bm{H}_i^k\rightarrow\bm{Y}_{i}^{k'}}]*\bm{Y}_{i}^{k'}\|_1}_{\mathcal{L}_\texttt{Rec}}+\underbrace{\|\mathcal{F}(\bm{z}_{T'}^{k'})*\bm{M}^{k'}-\mathcal{F}_{\texttt{init}}(\bm{z}_{T'}^{k'})*\bm{M}^{k'}\|_1}_{\mathcal{L}_{\texttt{Cst}}}\text{.}
\label{eq:3}
\end{equation}
Here, $\mathcal{L}_{\texttt{Rec}}$ denotes a reconstruction loss, and $\mathcal{L}_{\texttt{Cst}}$ represents a constraint term that ensures the consistency of the non-editable region $\bm{M}^{k'}$ of $\bm{\Gamma}^{k'}$.  $\mathcal{F}_{\texttt{init}}(\bm{z}^{k'}_{T'})=\mathcal{F}(\bm{z}^{k'}_{T'}).\texttt{detach()}$ indicates the initial features of $\bm{z}^{k'}_{T'}$ before conducting iterative latent region optimization. Our ADSR and SFSO strategies are employed during the above iterative latent region optimization process to overcome \textbf{Challenge \ref{chal:1}} and \textbf{Challenge \ref{chal:2}}, which are detailed in Section \ref{sec:adsr} and Section \ref{sec:sfso}, respectively.

\textbf{REMARK 2.} 
If $\eta^k=\texttt{Animation}$, then $k’>k$, which represents a cross-frame optimization paradigm, i.e., using the perturbed features $\mathcal{F}_{\texttt{ref}}(\bm{z}^k_{T'})[\bm{\Pi}_{\bm{H}_i^k\rightarrow\bm{Y}_{i}^{k'}}]$  to guide the denoising process of $\bm{z}^{k’}_{T}$ of the new frame $\bm{\Gamma}^{k’}$.
If $\eta^k=\texttt{Editing}$, $k’=k$, which can be seen as self-guided optimization, i.e., using the detached features $\mathcal{F}_{\texttt{ref}}(\bm{z}^k_{T'})[\bm{\Pi}_{\bm{H}_i^k\rightarrow\bm{Y}_{i}^{k'}}]$  of $\bm{\Gamma}^k$ to guide the re-denoising of $\bm{z}^k_{T}$.


\subsubsection{Adaptive Distribution Self-Rectification}
\label{sec:adsr}
We propose a simple-yet-effective strategy, Adaptive Distribution Self-Rectification (ADSR), to address the latent distribution drift issue caused by  cumulative perturbations---\textbf{Challenge~\ref{chal:1}}---as provided in \textbf{Proposition \ref{prop:dsclo}}.

\begin{proposition}[Adaptive Distribution Self-Rectification]
\label{prop:dsclo}
Suppose users apply drag-style operations to the frame $\bm{\Gamma}_k$. The statistics $\bar{\bm{\mu}}_{T'}$ and $\bar{\bm{\sigma}}_{T'}$ of the preceding neighboring latent embeddings $\{\bm{z}_{T'}^i\}_{i=k'-L_n-1:k'-1}$ of $\bm{\Gamma}_k$ are recorded, where $\bar{\bm{\mu}}_{T'}$ and $\bar{\bm{\sigma}}_{T'}$ are the mean and standard deviation. We propose using $\bar{\bm{\mu}}_{T'}$ and $\bar{\bm{\sigma}}_{T'}$ to rectify the distribution of $\bm{z}^{k'}_{T'}$ after each optimization iteration:  
\begin{equation}
\hat{\bm{z}}_{T'}^{k'} = \frac{\texttt{Iter\_optim}(\bm{z}_{T'}^{k'},\bm{U}^k) - \bm{\mu}^k_{T'}}{\bm{\sigma}^{k'}_{T'}} * \bar{\bm{\sigma}}_{T'} + \bar{\bm{\mu}}_{T'} \textit{,}
\label{eq:4}
\end{equation}
\end{proposition}
where $\texttt{Iter\_optim}(\cdot)$ denotes an iteration of the latent optimization, and $\bar{\bm{\mu}}_{T'}$$/$$\bm{\mu}_{T'}^{k'}$ and $\bar{\bm{\sigma}}_{T'}$$/$$\bm{\sigma}_{T'}^{k'}$ denotes the mean and standard deviation of $\{\bm{z}_{T'}^i\}_{i=k'-L_n-1:k'-1}$$/$$\bm{z}^{k'}_{T'}$. As exemplified in Figure \ref{fig:2} (a), our ADSR can effectively suppress the distribution drift of latent embeddings, while significantly improving video quality and preventing undesired changes in object attributes during dragging. This aligns with the findings provided in Figure~\ref{fig:7}, showing that ADSR consistently improves model performance across the evaluation metrics ObjMC, DAI, FVD, and FID. For more details, please refer to Section \ref{sec:exp}.

\subsubsection{Spatial-Frequency Selective Optimization}
\label{sec:sfso}
We design a Spatial-Frequency Selective Optimization (SFSO) mechanism to overcome \textbf{Challenge~\ref{chal:2}}. It fully exploits the information of context frames while relieving their interference via conducting information selection in both frequency and spatial domains during iterative latent region optimization.

High-frequency information—though capturing finer visual information—tends to mislead VDMs to produce unnatural results, as it carries more noise perturbations \cite{fan2019brief,li2020wavelet}; by contrast, low-frequency information—while more robust—lacks sufficient fine-grained visual details. We argue that it is crucial to harness the strengths of both high- and low-frequency information while alleviating their inherent limitations during the drag-oriented optimization process. We therefore propose a Switchable Frequency-domain Selection (SFS) strategy in \textbf{Proposition \ref{prop:sfs}}.

\begin{proposition}[Switchable Frequency-domain Selection]
\label{prop:sfs} 
Let $\{l_i\}_{i=1:L}$ represent the layers of the DiT denoiser that are used to construct reference features, and let $\bm{X}^k_{l_i}$ denote the input features of the layer $l_i$.  
SFS is applied to the self-attention of the layer $\{l_i\}_{i=1:L}$ to build reference features with switchable frequency components in each iteration of the latent region optimization process: 
\begin{equation}
\bm{Q}^k_{l_i},\bm{K}^k_{l_i},\bm{V}^k_{l_i}=\emph{\texttt{Linear\_projector}}(\bm{X}^k_{l_i})\text{,}
\label{eq:5}
\end{equation}
\begin{equation}
\bm{\bar{K}}^k_{l_i}=\emph{\texttt{Concat}}(\{\bm{K}^j_{l_i}\}_{j=0:k-1},\bm{K}^k_{l_i}),\bm{\bar{V}}^k_{l_i}=\emph{\texttt{Concat}}(\{\bm{V}^j_{l_i}\}_{j=0:k-1},\bm{V}^k_{l_i})\text{,}
\label{eq:6}
\end{equation}
\begin{equation}
\{\bm{\bar{K}}^k_{l_i},\bm{\bar{V}}^k_{l_i}\}=\emph{\texttt{IFFT}}(\emph{\texttt{Butterw}}(\emph{\texttt{FFT}}(\{\bm{\bar{K}}^k_{l_i},\bm{\bar{V}}^k_{l_i}\}),\omega=\emph{\texttt{Random}}(\omega_1,...,\omega_N)))\text{,}
\label{eq:8}
\end{equation}
\begin{equation}
\bm{\bar{X}}^k_{l_i}=\emph{\texttt{self-attention}}(\bm{Q}^k_{l_i}, \bm{\bar{K}}^k_{l_i}, \bm{\bar{V}}^k_{l_i})\text{.}
\label{eq:9}
\end{equation}
\end{proposition}
\emph{Here, $\{\bm{K}_{l_i}^j\}_{j=0:k-1}$ and 
$\{\bm{V}_{l_i}^j\}_{j=0:k-1}$ denote cached keys and values, $\bm{\bar{X}}^k_{l_i}$ denotes the extracted reference features of the layer $l_i$,
$\emph{\texttt{Butterw}}(\cdot \mid \omega)$ represents the Butterworth filter with the cutoff frequency $\omega$ randomly selected from $\{\omega_i\}_{i=1:N}$, and $\emph{\texttt{FFT}}(\cdot)$ and $\emph{\texttt{IFFT}}(\cdot)$ represent the 2D Fourier transform and 2D inverse Fourier transform.}

By using SFS strategy, in each iteration, the information of different frequencies can be propagated to the latent embeddings $\bm{z}^{k'}_{T'}$ of $\bm{\Gamma}^{k'}$ by the reconstruction loss $\mathcal{L}_{\texttt{Rec}}$, thus fully exploiting information from context frames, while preventing high-frequency information from dominating the drag process and inducing artifacts in generated frames.

In \textbf{Proposition \ref{prop:css}}, we also design a Criticality-driven Spatial-domain Selection (CSS) strategy to prevent over-optimization of the background within editable region, which is beneficial for further reducing unnatural content. 
\begin{proposition}[Criticality-driven Spatial-domain Selection]
\label{prop:css}
We selectively back-propagate gradients in spatial domain, avoiding the drag process undesirably affecting the background:
\begin{equation}
  \bm{z}_{T'}^{k'} \longleftarrow\bm{z}_{T'}^{k'} - \bm{G}^{k'}\frac{\partial\mathcal{L}_{\texttt{Tot}}}{\partial\bm{z}_{T'}^{k'}}
\label{eq:10}
\end{equation}
\end{proposition}
where $\bm{G}^{k'}$ is a Gaussian filtering map that decays w.r.t. the distance to the center point $(x_c, y_c)$ of the edited region
\begin{equation}
\bm{G}^{k'}[x,y] = \exp
\left[
    - \left(
        \frac{(x - x_c)^2}{2\sigma_x^2} + \frac{(y - y_c)^2}{2\sigma_y^2}
    \right)
\right], \,\,s.t., \,\,\sigma_x = \frac{W}{2} * \alpha \text{ and } \sigma_y = \frac{H}{2} * \alpha,
\label{eq:11}
\end{equation}
$W$ and $H$ are the width and height of the handle region's minimum bounding rectangle, and $\alpha$ is a hyperparameter scaling the spread of the Gaussian and set as $1$. The use of SFS and CSS can further improve video quality, which is demonstrated by experiments given in the main paper and appendix.

\section{Experiments}
\label{sec:exp}
\vspace{-0.3cm}
Since REVEL is a new task, no existing approaches have been specifically designed to tackle it. We adapt two training-free methods, SG-I2V \cite{sgi2v} and DragVideo \cite{dragvideo}, to the REVEL task for comparison. Please refer to Section \ref{sec:setup} of the appendix for details about our experimental setup, including implementation details, evaluation metrics, and compared baselines.

\begin{figure}[t!]
    \centering
    \includegraphics[width=0.9\linewidth]{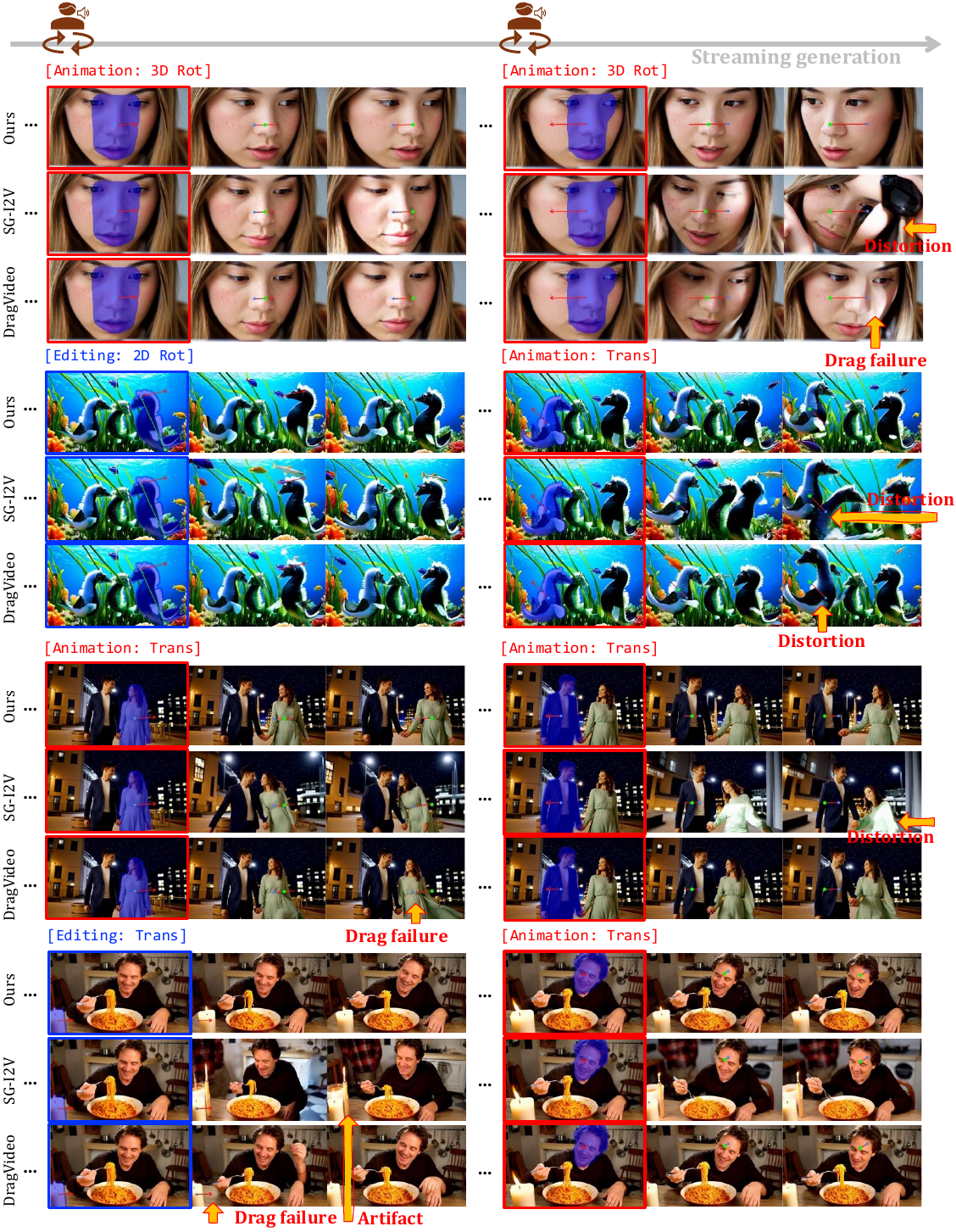}
    \vspace{-0.4cm}
    \caption{\textbf{Visualization results achieved by our DragStream on REVEL.} Note that \texttt{Editing} produces only one video frame, but we insert an extra subsequent frame to maintain layout consistency with \texttt{Animation}.}
    \label{fig:4}
\vspace{-0.5cm}
\end{figure}

\vspace{-0.3cm}

\subsection{Main Results}
\vspace{-0.3cm}
\textbf{Visualization Results.} The visualization results achieved by our method are shown in Figure~\ref{fig:4}. Compared to SG-I2V and DragVideo, our DragStream produces obviously more natural and higher-quality streaming drag-style video manipulation results. For instance, it better preserves object appearance and structure, while exhibiting fewer visual distortions, artifacts, and drag failures. These results validate the effectiveness of our method in addressing the REVEL task. More visualization results achieved by our DragStream are provide in the appendix; for details, please refer to Section~\ref{sec:morevis}. 

\begin{figure*}[t]
    \centering
    \begin{minipage}{0.92\linewidth}
        \centering
        \includegraphics[width=\linewidth]{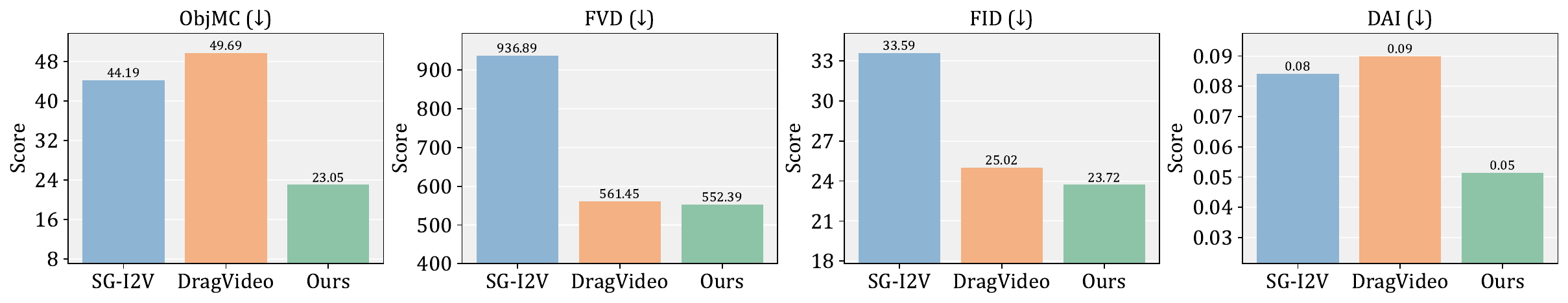}
        \vspace{-0.7cm}
        \caption{\textbf{Quantitative performance achieved by our method in terms of ObjMC, FVD, FID, and DAI.} ``$\downarrow$’’ indicates that lower values correspond to better performance.}
        \label{fig:5}
    \end{minipage}
    
    
    \begin{minipage}{0.93\linewidth}
        \centering
        \includegraphics[width=\linewidth]{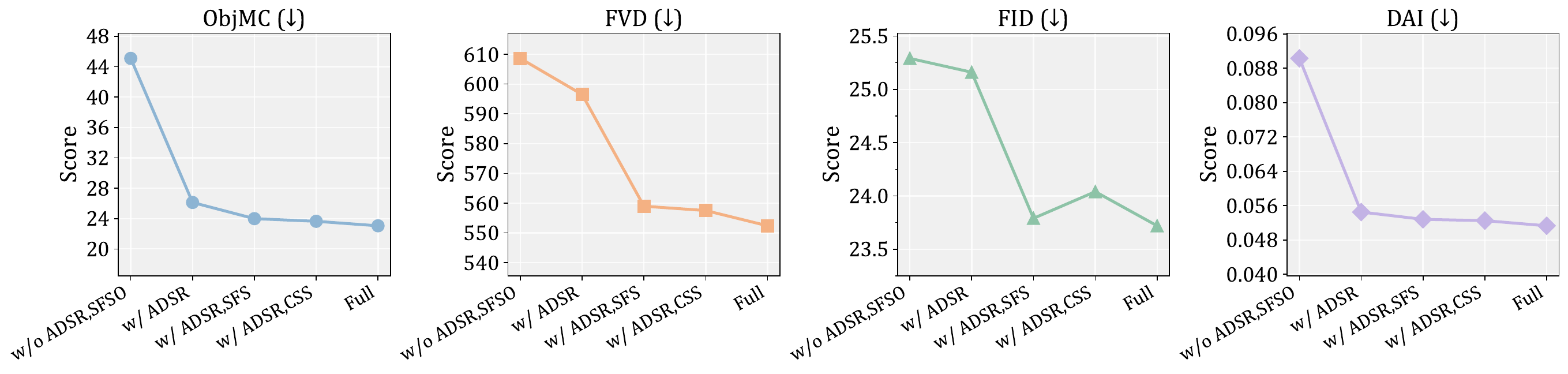}
        \vspace{-0.7cm}
        \caption{\textbf{Ablation study on the key components of our DragStream.}}
        \label{fig:6}
    \end{minipage}
    
    
    \begin{minipage}{0.9\linewidth}
        \centering
        \includegraphics[width=\linewidth]{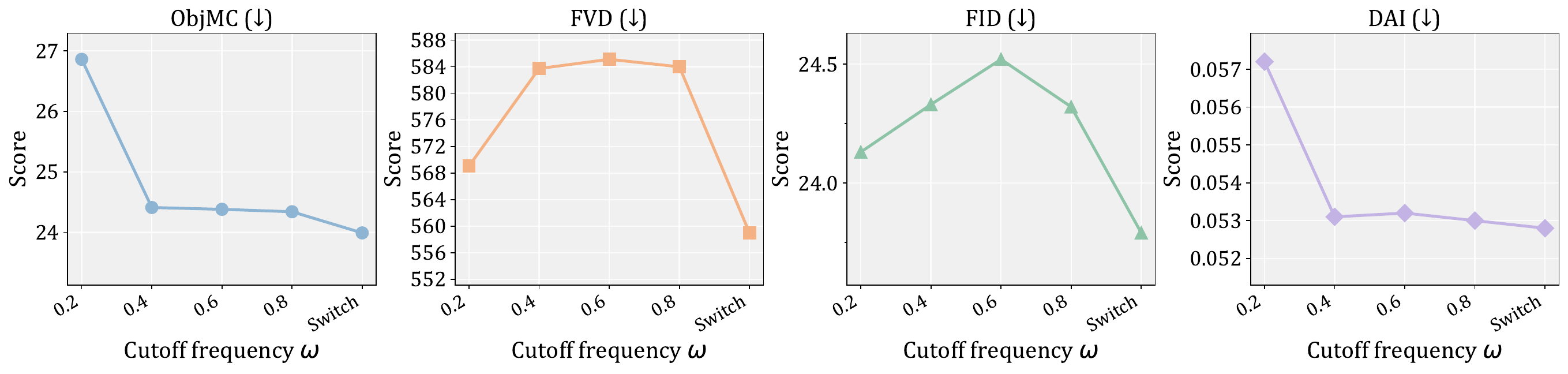}
        \vspace{-0.7cm}
        \caption{\textbf{Analysis on the influence of the cutoff frequency $\omega$.} ``Switch'' represents frequencies are switchable during the latent region optimization.}
        \label{fig:7}
    \end{minipage}
\vspace{-0.5cm}
\end{figure*}

\textbf{Quantitative Performance.} The quantitative results in Figure~\ref{fig:5} demonstrate that our DragStream consistently outperforms SG-I2V and DragVideo again. On one hand, the lowest FID and FVD scores indicate that our DragStream achieves higher video quality than SG-I2V and DragVideo. On the other hand, achieving the best ObjMC and DAI scores demonstrates that our DragStream approach realizes more precise object dragging, aligned with the findings shown in Figure~\ref{fig:4}.

\vspace{-0.3cm}

\subsection{Analysis}

\vspace{-0.3cm}

\mypar{Ablation Study.} In Figure~\ref{fig:6}, we conduct ablation study to investigate the influence of each component. The results indicate the full method achieves the best performance. Discarding SFSO (``w/ ADSR'') leads to significant performance degradation, while further removing ADSR (''w/o ADSR, SFSO'') results in an even greater decline. These results demonstrate the importance of the ADSR strategy and the SFSO mechanism. Similarly, using the full SFSO is better than using CSS or SFS alone.
We also analyze the influence of the cutoff frequency in Figure~\ref{fig:7}.
We can see that both small and large cutoff frequencies lead to performance drops. By contrast, our switchable frequency selection strategy achieves the best performance, as it fully exploits contextual information while mitigating high-frequency interference by preventing them from dominating the drag process.

\input{tables/efficiency}

\mypar{Runtime Analysis.} Table \ref{tab:efficiency} exhibits the runtime analysis of our DragStream approach. Our DragStream is based on an iterative optimization scheme. In the table, we investigate the influence of the iteration number $I$. We find that setting $I = 4$ already achieves satisfactory performance, achieving 23.05 ObjMC and 0.051 DAI, while incurring only 0.13s of additional runtime per frame compared with the baseline without DragStream (i.e., $I = 0$). Decreasing the iteration number---such as $I = 2$ or $3$---can further improve execution speed, while still maintaining acceptable drag-based manipulation performance, with ObjMC and DAI clearly outperforming those of the baseline (i.e., $I=0$). All the experiments on the table are conducted on a NVIDIA H20 GPU.

\begin{figure}[t!]
    \centering
    \includegraphics[width=0.9\linewidth]{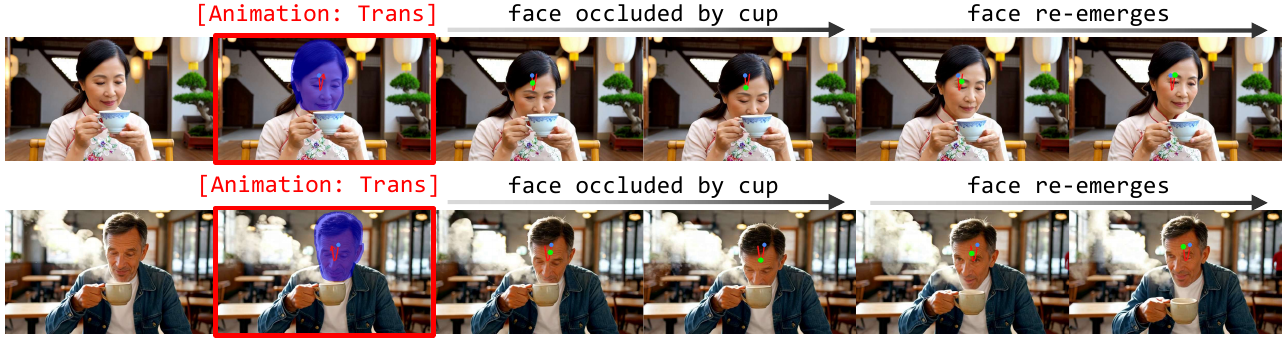}
    \vspace{-0.3cm}
    \caption{\textbf{Streaming drag with object occlusion and Re-emergence.}}
    \vspace{-0.4cm}
    \label{fig:occlusion}
\end{figure}

\begin{figure}[h!]
    \centering
    \includegraphics[width=0.9\linewidth]{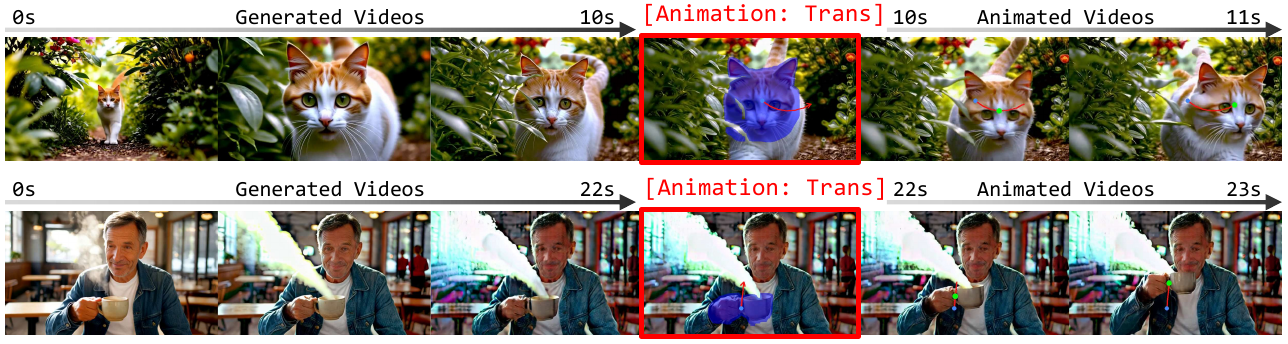}
    \vspace{-0.3cm}
    \caption{\textbf{Streaming drag in long video scenarios.}}
    \vspace{-0.4cm}
    \label{fig:drag_long_video}
\end{figure} 

\begin{figure}[t!]
    \centering
    \includegraphics[width=0.9\linewidth]{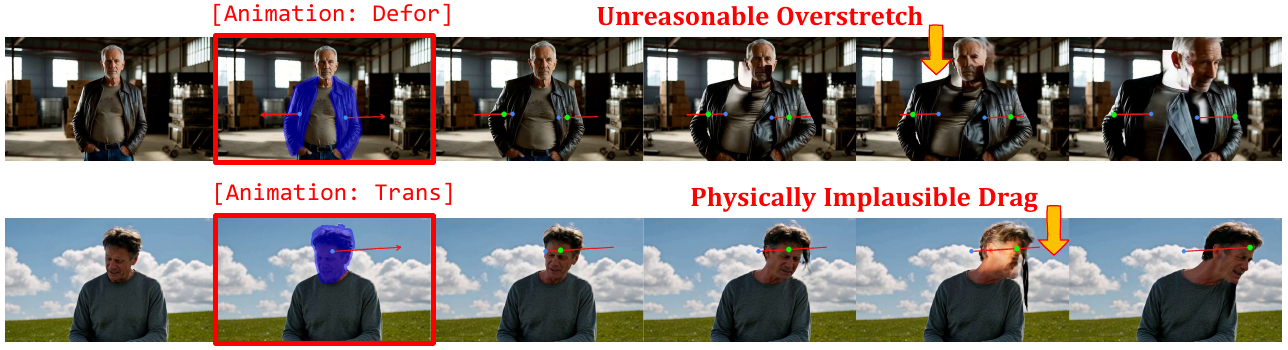}
    \vspace{-0.3cm}
    \caption{\textbf{Failure cases under unreasonable and physically implausible conditions.}}
    \vspace{-0.4cm}
    \label{fig:failure}
\end{figure}

\subsection{Complex Streaming Manipulation}
\label{sec:complex_scenes}

\vspace{-0.3cm}


\mypar{Occlusion and Re-emergence.}
In Figure \ref{fig:occlusion}, we also study our DragStream in the scenario of object occlusion and subsequent re-emergence. We find that our approach shows promising performance in this scenario and produce smooth video results. This is because VDMs are trained on massive amounts of data and thereby learns rich prior knowledge about object occlusion and scene transition.

\vspace{-0.2cm}

\mypar{Streaming Drag in Long Video Generation.}
In Figure \ref{fig:drag_long_video}, we study the use of our DragStream for achieving streaming drag in long video generation. As shown in the figure, despite that accumulated errors remain a challenging issue for current autoregressive VDMs, our method can still effectively realize drag-based manipulation. For more results, please refer to Section \ref{sec:additional_long_results} of our appendix.

\subsection{Failure Cases}
\vspace{-0.3cm}
We observe a failure case of our method. As shown in Figure \ref{fig:failure}, our method fails to realize high-quality manipulation under highly unreasonable and physically implausible conditions, as such manipulation instructions severely conflict with prior knowledge learned by VDMs in large-scale data.

\vspace{-0.3cm}

\section{Conclusion}
\vspace{-0.3cm}
We propose \textbf{stReaming drag-oriEnted interactiVe vidEo manipuLation (REVEL)}, a new task that aims to allow users to achieve streaming, drag-style control over the outputs of autoregressive VDMs. To solve REVEL, we propose a training-free approach, \textbf{DragStream}, which employs an Adaptive Distribution Self-Rectification (ADSR) strategy and design a Spatial-Frequency Selective Optimization (SFSO) mechanism. ADSR effectively constrains  the drift of latent embeddings by leveraging neighboring frames' statistics, while SFSO fully exploits contextual information while mitigating its interference via selectively propagating visual cues along generation in spatial and frequency domains. These two strategies enable our method to achieve superior performance on REVEL and allow seamless integration into existing autoregressive VDMs. We hope this work will inspire more excellent solutions to address the streaming drag-style video manipulation problem.

\section*{Acknowledgments}
This research is supported by the National Research Foundation, Singapore under its AI Singapore Programme (AISG Award No: AISG3-RP-2022-030) and NRF-NRFI10-2024-0004.

\bibliography{iclr2026_conference}
\bibliographystyle{iclr2026_conference}

\clearpage

\end{document}

%% file: math_commands.tex
\usepackage{amsmath,amsfonts,bm}

\def\eqref#1{equation~\ref{#1}}

\def\1{\bm{1}}

\DeclareMathAlphabet{\mathsfit}{\encodingdefault}{\sfdefault}{m}{sl}
\SetMathAlphabet{\mathsfit}{bold}{\encodingdefault}{\sfdefault}{bx}{n}



%% file: tables/efficiency.tex
\begin{wraptable}{r}{0.48\textwidth}
    \centering
    \small
    \captionsetup{font=scriptsize}

    \vspace{-4mm}
    \caption{\textbf{Runtime analysis of our DragStream approach.} In the table, ``RF'' denotes runtime per frame, and $I$ indicates the number of iterations of drag-oriented latent optimization.}
    \vspace{-3mm}
    \begin{tabular}
        {c | c c c}
        \toprule
        \textbf{Experiments}               & \textbf{RF}  & \textbf{ObjMC} ($\downarrow$) & \textbf{DAI} ($\downarrow$) \\
        \midrule
        $I=0$      & \textbf{0.17s} & 90.39                & 0.133
        \\
        \midrule
        $I=2$                 & 0.24s          & 27.67                & 0.054           \\
        $I=3$                 & 0.27s          & 24.55                & 0.053           \\
        \midrule
        \textbf{$I=4$ (Ours)} & 0.30s          & \textbf{23.05}       & \textbf{0.051}  \\
        \bottomrule
    \end{tabular}
    \label{tab:efficiency}
    \vspace{-4mm}
\end{wraptable}